%% file: main.tex
\documentclass[10pt,twocolumn,letterpaper]{article}

\usepackage[pagenumbers]{iccv} 

\input{preamble}

\definecolor{iccvblue}{rgb}{0.21,0.49,0.74}
\usepackage[pagebackref,breaklinks,colorlinks,allcolors=iccvblue]{hyperref}

\usepackage{soul}

\def\paperID{6639} 
\def\confName{ICCV}
\def\confYear{2025}

\title{The Devil is in the Spurious Correlations: \\ Boosting Moment Retrieval with Dynamic Learning}

\author{Xinyang Zhou$^{1\star} $ \quad Fanyue Wei$^{2\star}$ \quad Lixin Duan$^1$ \quad Angela Yao$^2$ \quad Wen Li$^1$
\vspace{3pt}\\
\normalsize$^1$University of Electronic Science and Technology of China \quad $^2$National University of Singapore\\
\tt\small \{x.y.zhou2020, lxduan, liwenbnu\}@gmail.com \\
\tt\small \{wfanyue, ayao\}@ comp.nus.edu.sg}

\begin{document}
\maketitle
\footnotetext{$^{\star}$Equal contribution.}

\input{sec/0_abstract}




\input{sec/1_intro}

\input{sec/2_related_work}
\input{sec/3_method}
\input{sec/4_exp}

\section{Conclusion}
Although existing transformer-based approaches have demonstrated remarkable performance, they still struggle with spurious correlation, which overly associates the textual query with the background frames rather than the target moment. To address this issue in moment retrieval, we introduce a novel dynamic learning method with two strategies. First, we propose a novel video synthesis strategy that constructs a dynamic context for the relevant moment. This synthesis strategy enables our model to attend to the target moment corresponding to the query across various dynamic video contexts. Second, we enhance the representation by learning temporal dynamics aligned with texts. In addition to visual features, text queries are aligned with dynamic representations, encouraging our model to establish a non-spurious correlation between the query-related moment and its context. Extensive experiments and detailed ablation studies on both the QVHighlights and Charades-STA benchmarks validate the effectiveness and generalization of the proposed \emph{TD-DETR}, demonstrating significant alleviation of spurious correlation and superior performance in moment retrieval.
{
    \small
    \bibliographystyle{ieeenat_fullname}
    \bibliography{main}
}

\input{sec/X_supple}

\end{document}

%% file: preamble.tex
\usepackage{multirow}
\usepackage{boldline}
\usepackage{pifont}
\usepackage{amssymb}
\usepackage{bm}

\usepackage{threeparttable}

\definecolor{Bpurple}{HTML}{BF40BF}

\usepackage{colortbl}

\newcommand{\red}[1]{{\color{red}#1}}

\newcommand{\blue}[1]{{\color{blue}#1}}


%% file: sec/0_abstract.tex
\vspace{-0.4cm}
\begin{abstract}
Given a textual query along with a corresponding video, the objective of moment retrieval aims to localize the moments relevant to the query within the video. While commendable results have been demonstrated by existing transformer-based approaches, predicting the accurate temporal span of the target moment is still a major challenge. This paper reveals that 
a crucial reason stems from the spurious correlation between the text query and the moment context. Namely, the model makes predictions by overly associating queries with background frames rather than distinguishing target moments.
To address this issue, we propose a dynamic learning approach for moment retrieval, where two strategies are designed to mitigate the spurious correlation. First, we introduce a novel video synthesis approach to construct a dynamic context for the queried moment, enabling the model to attend to the target moment of the corresponding query across dynamic backgrounds. Second, to alleviate the over-association with backgrounds, we enhance representations temporally by incorporating text-dynamics interaction, which encourages the model to align text with target moments through complementary dynamic representations.
With the proposed method, our model significantly alleviates the spurious correlation issue in moment retrieval and establishes new state-of-the-art performance on two popular benchmarks, \ie, QVHighlights and Charades-STA. In addition, detailed ablation studies and evaluations across different architectures demonstrate the generalization and effectiveness of the proposed strategies. Our code will be publicly available.


\end{abstract}

%% file: sec/1_intro.tex
\section{Introduction}
\label{sec:intro}

Videos 
are 
a ubiquitous media format, but 
it can be time-consuming to browse through videos 
to localize specific moments. 
Using 
text to retrieve corresponding moments within a lengthy video\cite{zhang2023temporal, anne2017localizing} plays an important role 
for entertainment, search, \etc. 

Existing moment retrieval approaches~\cite{lei2021detecting, moon2023query, yang2024task, lee2025bam, jung2024bmdetr} leverage the DETR~\cite{carion2020end} detection transformer architecture to fuse text query into video representation.   
They show remarkable results in detecting the target moment by text-video alignment, but accurately predicting the temporal span remains a significant challenge. The learned video representations~\cite{lei2021detecting, lee2025bam} are aligned with the semantics of the text, thus generally biased to subjects or spatial objects. Consequently, the semantic elements in background frames are overly associated with textual queries. 
Such inappropriate association between irrelevant contexts and labels is the manifestation of \emph{Spurious correlation}~\cite{agarwal2020towards_spurious, wang2024effect_spurious}, which are under-explored in moment retrieval tasks. 

This work identifies such 
\emph{spurious correlations} between the text queries and moment visual context as the root cause for this performance gap in moment retrieval.  Specifically, learned SOTA moment retrievers~\cite {lee2025bam, moon2023query, lei2021detecting} tend to associate text queries with the background frames rather than distinguishing the target moment.  Figure~\ref{fig:intro_overview} illustrates how such spurious issues lead to a sub-optimal performance for moment retrieval. 
For example, the SOTA method,  BAM-DETR~\cite{lee2025bam}, predicts a similar temporal span for an \texttt{`The blonde woman works out in a gym with red lighting.'} in both the original video and an altered version where the target moment is masked out.  In the masked case, the segment near the mask is a more reasonable output but the baseline methods~\cite{moon2023query, lee2025bam, yang2024task, moon2023correlation} predict the masked segment from background frames with the learned spurious correlation, which is the most unrelated output to the text query. 

To address the spurious issues, we propose a dynamic learning method with two novel strategies 
to mitigate spurious correlation. 
First, we propose a synthesis strategy that dynamically contextualizes the target moments for retrieval. We select similar video pairs and augment the samples by synthesizing target moments within similar contexts. This encourages the model to distinguish target moments from queries across more dynamic contexts, alleviating the over-association with background frames.
Second, we propose to enhance video-text representations to reduce spurious dependencies on background frames by aligning text queries with temporal dynamics. To achieve this, we design a simple yet efficient temporal tokenizer to extract dynamic representations. Additionally, a text-dynamics interaction module is introduced to align dynamics with text queries, enabling the model to establish a stand-up correlation with complementary dynamic representations while mitigating background bias.

Experiments on two challenging benchmarks, \ie, QVHighlights and Charades-STA, show that our method outperforms existing methods by clear margins. Besides, \emph{spurious correlation} in moment retrieval is validated to be largely alleviated by spurious evaluation. The detailed ablation analyses demonstrate the effectiveness of the proposed strategies to resolve the spurious issues.

In conclusion, our contributions to this work are summarized as follows:
\begin{itemize}
    \item To the best of our knowledge, we are the first to investigate the spurious correlation in moment retrieval.
    \item We propose a dynamic learning framework for moment retrieval that mitigates spurious correlations by dynamically contextualizing target moments through novel video synthesis and enhancing representations with text-dynamics alignment.
    \item The proposed method achieves SOTA performance across all benchmarks. Besides, our model also provides a strong interpretation of alleviating \emph{spurious correlations}.
\end{itemize}

%% file: sec/2_related_work.tex
\section{Related Work}
\label{sec:Related Work}

\subsection{Moment Retrieval} 
Moment Retrieval aims to predict a temporal span within a video corresponding to a given natural language query~\cite{zhang2023temporal}. 
Early methods 
had two stages: first, sampling candidate moments as proposals, and then scoring the proposals to obtain the final predictions~\cite{gao2017tall, anne2017localizing, liu2018cross, liu2018attentive, ge2019mac}. However, more recent 
approaches are one-shot~\cite{zhang2019man, yuan2019semantic, zhang2020learning, zeng2020dense, zhao2017ssn, lei2020tvr, regneri2013tacos, xu2023boundary} to predict the temporal span directly. Simultaneously, hybrid frameworks that integrate transformers for feature learning and fusion with CNNs for prediction have been widely studied. SnAG~\cite{mu2024snag} explored the impact of cross-modal fusion on the scalability of video grounding models and proposed an efficient training scheme.


Although many moment retrieval methods have achieved remarkable performance, the majority primarily focus on architectural design, such as exploring new networks~\cite{zhang2019man, yuan2019semantic}, enhancing video-text alignment\cite{lei2020tvr}, or refining attention mechanisms\cite{mu2024snag}. Generally, these methods focus on how to fuse text and video representations to improve the temporal span prediction but the overall performance are limited by the inefficient alignment architecture.


\begin{figure*}[t]
    \centering
    \includegraphics[width=\linewidth]{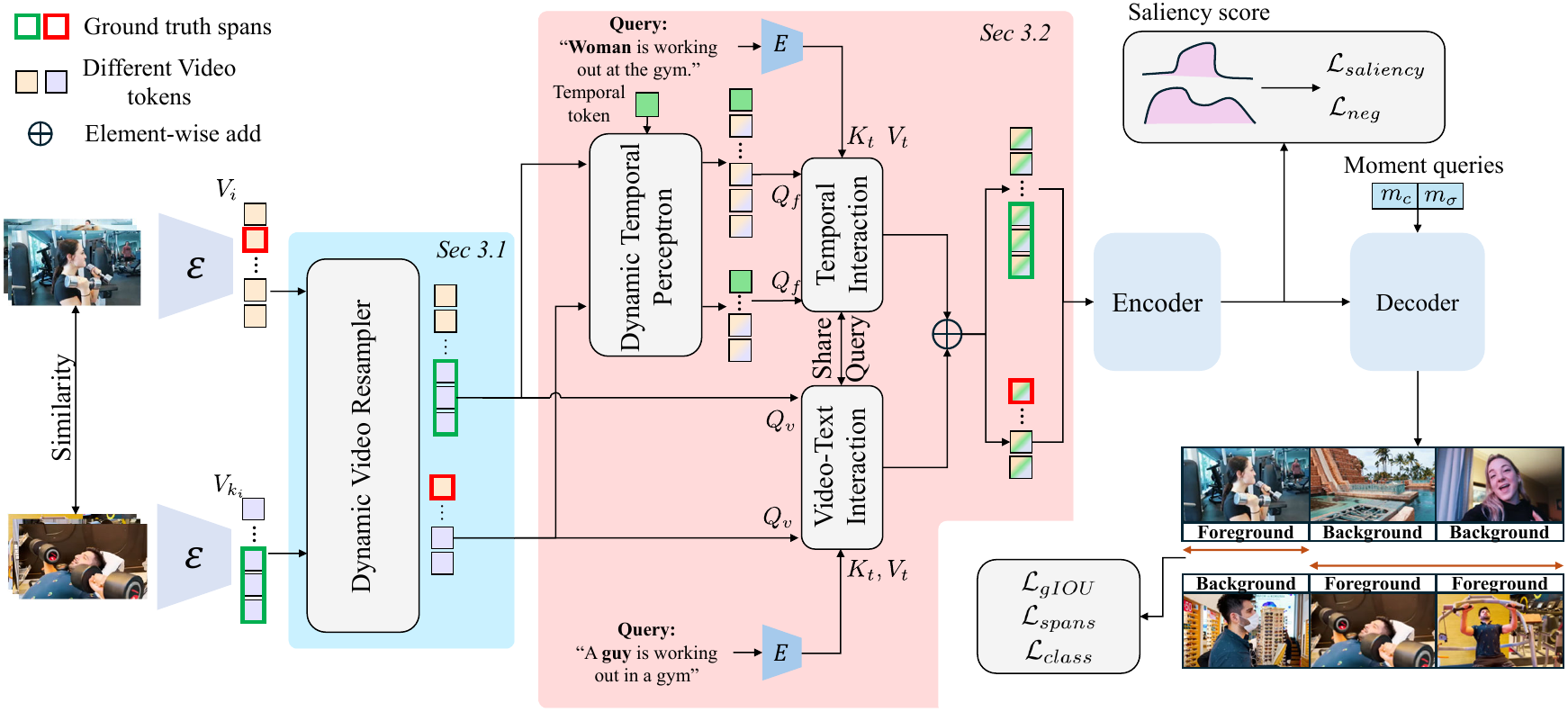}
    \caption{ Overview of the proposed \emph{TD-DETR}. Video pairs are sampled by their similarity and then forwarded into the Video Synthesizer to generate Dynamic Context (Section~\ref{sec::VSDC}), Then, the Dynamic Context, \ie, synthesized pairs, are enhanced by Temporal Dynamic Tokenization and interact with textual information (Section~\ref{sec::TDT}). Finally, the enhanced video tokens are sent to the transformer encoder-decoder with prediction heads to predict moments (Section~\ref{sec::network}).
    }
    \label{fig:method_overview}
\end{figure*}

\subsection{Detection Transformers}
In the past several years, the wide adoption of detection transformers (DETR)~\cite{carion2020end} in video moment retrieval has gained remarkable performance~\cite{jang2023eatr, lei2021detecting, liu2022umt, moon2023query, lee2025bam, yang2024task, xu2023mh-detr}. While introducing the DETR framework to the task of moment retrieval, it demonstrated the effectiveness on QVHighlights dataset~\cite{lei2021detecting}. 
Furthermore, DETR-based approaches have explored the relationship between moment retrieval and highlight detection, achieving significant advancements in both tasks. However, they still struggle with inaccurate span predictions~\cite{otani2020uncovering, zhang2023temporal}, limiting their overall reliability. To address this issue, Zhang et al.~\cite{zhang2023temporal} analyzes the distribution of target moments and highlights the problem of imbalance, while Otani et al.~\cite{otani2020uncovering} investigate the skewed distribution of query representations and expose the overly query-dependent phenomenon of existing models. Unlike previous studies, we reveal that \emph{spurious correlations} is the key cause of inaccurate span predictions, arising from the over-association between the text query and the background frames.

\subsection{Spurious Correlation}
\label{sec:spurious_correlation}
One reason for the poor generalization of vision algorithms is that they are prone to 
memorizing patterns or contextual cues
~\cite{ghosal2024vision, beery2018recognition, sagawa2019distributionally, geirhos2018imagenet, goel2020model, tu2020empirical}. These patterns or cues are often \emph{spurious correlations} - 
misleading heuristics 
of the training data 
correlated with the majority of examples but does not hold in general~\cite{ghosal2024vision}. While numerous approaches have been proposed to address this challenge~\cite{wang2020identifying, wang2024effect, zhang2022correct, zhang2021videolt, otani2020uncovering, zhang2023temporal, zhang2023deep} in various domains, their impact on video understanding remains largely unexplored. For example, in image-based tasks, spurious correlations are often tied to spatial biases~\cite{agarwal2020towards_spurious, wang2024effect_spurious}, whereas in video-related tasks, such spatial biases are redundant in frames. This distinction is particularly crucial in retrieving moments from lengthy videos, where the goal is not merely to recognize objects but to precisely localize temporal spans with text queries, making the study of spurious correlations in this context even more critical.

Despite in Moment Retrieval, Otani et al.~\cite{otani2020uncovering} investigated the biases introduced by text queries and the neglect of semantic information, while Zhang et al.\cite{zhang2023temporal} explores the distribution of start and end moments. However, the issue of spurious correlation between the text and background frames is a key challenge of moment retrieval. In this work, we focus on this point and develop a new method by learning dynamics to alleviate this problem.

%% file: sec/3_method.tex
\section{Methodology}
\label{sec::Methodology}
The objective of moment retrieval is to localize a temporal span that semantically corresponds to a given text query. Consider a video represented by $L$ vision tokens, denoted by $V=\{v_1, v_2, \dots, v_L\}$, along with a natural language description of $W$ words, as $\{q_1, q_2, \dots, q_W\}$. 
A moment retrieval model 
predict a temporal span $m$ with centre $m_c$ and duration $m_{\sigma}$ which is relevant to the textual query, and token-wise saliency scores $\{s_1, s_2, \dots, s_L\}$ of the text. 


As shown in Figure~\ref{fig:method_overview}, we propose a novel framework to learn \textit{\textbf{T}}emporal \textit{\textbf{D}}ynamics utilizing \textit{\textbf{DE}}tection \textit{\textbf{TR}}ansformer (\textbf{TD-DETR}). The video pairs are sampled by similarity and then forwarded into the video synthesizer to generate dynamic contexts. Then, the representations of synthesized pairs are enhanced through temporal dynamic interaction with texts. Finally, dynamic-enhanced representations are fed into the transformer encoder-decoder with prediction heads to predict target moments.

\subsection{Video Synthesizer for Dynamic Context}
\label{sec::VSDC}
We introduce a novel video synthesizer to infuse the target moments with dynamic context for moment retrieval. Spurious correlations often stem from 
linking the moment's context 
to the text query.  We 
address the issue by synthesizing new samples for the target moments with more dynamic contextual variations. 


\noindent \textbf{Spurious Pair Selection.} In order to synthesize a dynamic context aligned with the target moment, we construct a spurious pair of the target moment in a given video $V$. 
To ensure the challenge and rationality of the synthesized video, we select a video which is similar to $V$ contextually.

In a training batch of $N$ video samples, we sample the most similar video $\{V_{k}\}$ for every $\{V_i\}$ where $k \in [1, N], k \neq i$, for each $i \in [1, N]$ to construct a spurious video pair $p_{i}$.
For any videos $V_j$ and $V_l$ from a batch, we employ cosine similarity to model the similarity relationship between videos as follows:
\begin{equation}
    s_{j,l} = \frac{1}{L_j \cdot L_l} \sum_{p=0}^{L_j} \sum_{q=0, p\neq q}^{L_l} \frac{v{^j_p}^T\cdot v^l_q}{\Vert v^j_p \Vert\cdot\Vert v^l_q \Vert} \quad j,l \in [1, N].
\label{eq:similarity}
\end{equation}

In Eq.~\ref{eq:similarity}, we denote $v^j_q$ and $v^l_p$ $q$-th clip of $V_j$ and $q$-th clip of $V_l$ , $L_j$ and $L_l$ for the length of $V_j $ and $V_l$, $a_{j,l}$ for the average similarity between $V_{j}$ and $V_{l}$. We select the most similar video $V_{k}$ for every video $V_i$ by $k = \mathop{\arg\max}_{k\in [0, N]}s_{[i,:\setminus \{i\}]}$ in the batch. 

\begin{figure}[t]
    \centering
    \includegraphics[width=\linewidth]{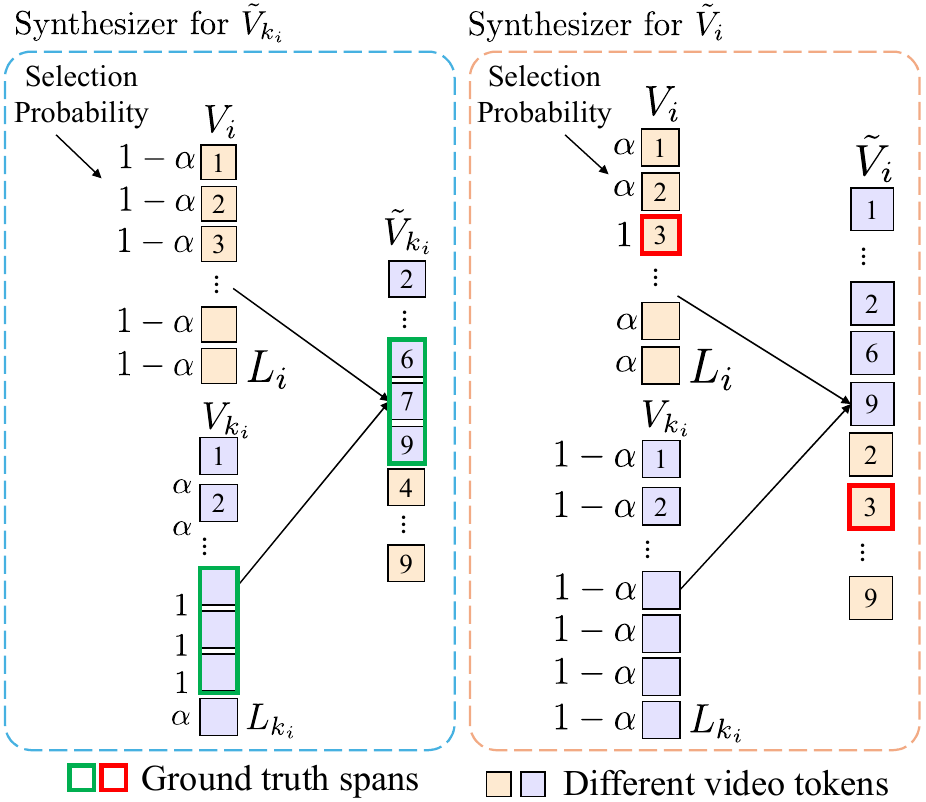}
    \caption{Illustration of Video Synthesis with Dynamic Context. The numbers in the boxes represent the token indices, indicating their sequential order. Vision tokens are selected based on their selection probabilities and concatenated while preserving the completeness of the ground truth spans, \ie, selection probabilities of ground truth spans are set to 1. The synthesized videos maintain their original length with a random bias.
    }
    \label{fig:method_DVS}
\end{figure}

\noindent \textbf{Video Synthesis with Dynamic Context.} After obtaining the spurious pairs, we synthesize a new video $\tilde{V_{i}}$ using the target moment $m$ and the spurious pair $p_{i}$, where $p_i = \{V_i, V_{k}\}$. We then composite $\{V_i, V_{k}\}$ and dynamically refine the ground truth of the target moment $m_{i}$.
Illustrated by Figure~\ref{fig:method_DVS}, given video $V_i$ with $L_i$ tokens and $V_{k}$ with $L_{k}$ tokens, on the one hand, we expect the sample from $V_i$ and $V_{k}$ with the completeness of target moment $m_i$ of $V_i$. We first sample from $V_i$ without ground truth tokens $\text{NG}\in[0, L_i] \setminus m_i$ with sampling ratio $\alpha$, \ie, every token in $\text{NG}$ has the same ratio $\alpha$ to be selected. We sample from the paired $V_{k}$ with sampling ratio $1-\alpha$, since $V_{k}$ is irrelevant to ground truth $\text{GT}_i$. 

Finally, we concatenate the two sets of sampled video tokens into $\tilde{V}_i$. On the other hand, we have the same process, except now we focus on $V_{k}$ and $\text{GT}_{k}$ with sampling ratio $\alpha$ and ratio $1-\alpha$ to sample from the whole $V_i$ and concatenate sampled tokens into $\tilde{V}_{k}$.

\subsection{Dynamics Enhancement}
\label{sec::TDT}

The attention of DETR-like architecture associates text queries with visual representations. Unfortunately, this tends to emphasize background frames, thereby exacerbating the \emph{spurious correlation}. To this end, besides the dynamic context synthesis, we introduce a dynamics enhancement module to encourage our model to align text queries with temporal dynamic representations. Thus, the model considers not only the association of background frames but also background-independent dynamics. This alignment enables our model to establish a stand-up correlation between the query-related moment and its context.

\noindent \textbf{Dynamic Tokenizer.}
To model the dynamic temporality, we introduce a simple yet effective strategy to tokenize the temporal dynamics, which is nearly cost-free to obtain temporal dynamic representations.

Given spurious pair $p_i=\{\tilde{V}_i, \tilde{V}_{k}\}$, the temporal dynamic tokenizer processes all video pairs in the same manner. Then, we use $\tilde{V}_i$ to explain the processing steps without loss of generality.

First, we concatenate a learnable token to the start of the video as follows:
\begin{equation}
    \tilde{V}_i=\{st, \tilde{v}_1, \tilde{v}_2, \dots, \tilde{v}_{L}\},
\end{equation}
where $st$ is a learnable token that captures the start signal of the video, preventing the loss of information from the first clip in the video representations. We then employ an element-wise difference to model the temporal dynamics.
\begin{equation}
    T=\{\tilde{v}_1 - st, \tilde{v}_2-\tilde{v}_1, \tilde{v}_3-\tilde{v}_2, \dots, \tilde{v}_{{L_i}} - \tilde{v}_{{L_i}-1}\},
\label{eq:dynamic-repre}
\end{equation}
where $T$ is a learnable dynamic representation, it learns and focuses on the dynamic information in nearby video clips.

\noindent \textbf{Text-Dynamics Interaction.}
After obtaining dynamic tokens $T$ in Eq.~\ref{eq:dynamic-repre} from Dynamic Tokenizer, we utilize cross-attention for both text-vision and text-dynamic representations with the same text query to mitigate the over-emphasis on video backgrounds. 


In detail, the \textit{query} for each cross-attention layer is prepared by linear projection of the video and temporal clips as follows:
\begin{equation}
    Q_{\delta} = \left[p_{\delta}(\delta),...,p_{\delta}(\delta_{L_i})\right], \delta\in \{\tilde{V}_i,T\}.
\end{equation}
The \textit{key} and \textit{value} are computed with the query text representation as $K_q=\left[p_k(q_1),...,p_k(q_W)\right]$ and $V_q=\left[p_v(q_1),...,p_v(q_L)\right]$, $p_{\delta}$, $p_k$, $p_v$ are linear projection layers for dynamic-enhanced \textit{query}, temporal \textit{query}, \textit{key} and \textit{value}. Then the cross-attention layer operates as follows:
\begin{equation}
    \delta'=\text{softmax}(\frac{Q_{\delta}K_q^T}{\sqrt{d}})V_q, \delta\in \{\tilde{V}_i,T\}, \delta'\in \{\tilde{V}'_i,T'\},
\end{equation}
where $d$ is the dimension of the projected \textit{key}, \textit{value} and \textit{query}. To emphasize the learned temporal information, we utilize the weighted element-wise addition to inject text-guided temporal representation into text-guided dynamic-enhanced representation as follows:
\begin{equation}
    \tilde{V}_i' = \beta\cdot \tilde{V}_i + (1-\beta)\cdot T',
\end{equation}
where $\beta$ is a hyper-parameter to adapt the addition ratio between $\tilde{V}'_i$ and $T'$.
Hence, from the spurious pair $p_i=\{\tilde{V}_i, \tilde{V}_{k}\}$, we obtain a new spurious pair $p_i'=\{\tilde{V}'_i, \tilde{V}'_{k}\}$.


\subsection{Network and Objectives} 
\label{sec::network}
Our network structure (see Figure~\ref{fig:method_overview}) follows previous work \cite{lei2021detecting, moon2023query}.  It features the synthesis of dynamic contextual videos and the enhancement of dynamic representations.

\noindent \textbf{Transformer encoder-decoder with prediction heads.} 
Given spurious pair $p_i'=\{\tilde{V}'_i, \tilde{V}'_{k}\}$, this module processes all video pairs in the same manner. Therefore, we use $\tilde{V}'_i$ to explain the processing steps without loss of generality.
The encoder consists of \( T \) stacked typical transformer encoder layers, as in \cite{moon2023query, lei2021detecting}, producing encoded representations \( E_{enc} \). Our decoder, following \cite{moon2023query}, also uses \( T \) stacked typical transformer decoder layers, along with \( N \) learnable moment queries representing the centre \( m_c \) and duration \( m_{\sigma} \). The decoder processes \( E_{enc} \) with the moment queries.

We adopt prediction heads following \cite{moon2023query}. A linear layer predicts saliency scores from the encoded representations, and another linear layer handles negative pairs. From the decoder outputs, a 3-layer MLP with ReLU predicts the normalized moment center and duration, while a linear layer with softmax predicts the foreground.

\noindent \textbf{Hungarian matching.}
Following~\cite{carion2020end, lei2021detecting, moon2023query}, we perform Hungarian matching between the two predictions and two corresponding labels respectively.
Give prediction $\hat{y}$ of $\tilde{V}'_i$ and corresponding to ground truth $y$, the optimal matching results between predictions and ground truths $\hat{\sigma}$ can be written as $\hat{\sigma} = \mathop{\arg\min}_{\sigma\in G_N}\sum_{i}^{N} \mathcal{C}_{\mathrm{match}}(y, \hat{y}_{\sigma(i)}),$
where $G$ is a permutation of predictions and ground truths pairs and $\mathcal{C}_{\mathrm{match}}$ is the matching cost.

\noindent \textbf{Loss Functions.}
We calculate the loss between $\hat{\sigma}$ and the ground truth $y$ corresponding to $\tilde{V}'_i$.
Following~\cite{lei2021detecting}, the $L_{1}$ loss $\mathcal{L}_{L_{1}}$ and the gIoU~\cite{rezatofighi2019giou} loss $\mathcal{L}_{gIoU}$ are used to measure the distance and overlapping between the predictions and 
a cross-entropy loss $\mathcal{L}_{cls}$ is used to measure classification.
For highlight detection, we also use three loss functions which are margin ranking loss $\mathcal{L}_{margin}$, rank-aware contrastive loss $\mathcal{L}_{margin}$ and negative loss $\mathcal{L}_{neg}$.

The overall loss is shown as follows with $\lambda_{*}$ as balancing coefficient:
    \begin{gather}
    \mathcal{L}_{moment} = \lambda_{L_{1}}\mathcal{L}_{L_{1}} + \lambda_{iou}\mathcal{L}_{gIoU} + \lambda_{cls}\mathcal{L}_{cls},\\
        \mathcal{L}_{hl}=\lambda_{margin}\mathcal{L}_{margin}+\lambda_{cont}\mathcal{L}_{cont}+\lambda_{neg}\mathcal{L}_{neg},\\
        \mathcal{L}_{total}=\mathcal{L}_{hl}+\mathcal{L}_{moment}.
    \end{gather}

%% file: sec/4_exp.tex
\section{Experiments}
\input{table/table_QVH_result}
\input{table/table_charades_results}

\subsection{Experimental Setup}
\label{sec:set_up}
\noindent \textbf{Datasets.} 
We evaluate our method on two widely-used benchmarks: QVHighlights~\cite{lei2021detecting} and Charades-STA~\cite{gao2017tall}, following the 
setup of prior works~\cite{lei2021detecting, moon2023query}. 
    \emph{QVHighlights} has over $10,000$ video-query pairs primarily from vlog and news content. 
    We use the splits 
    defined in Moment-DETR~\cite{lei2021detecting} and report 
    results for both the \textit{val} and \textit{test} splits. 
    \emph{Charades-STA}, 
    derived from the Charades dataset~\cite{sigurdsson2016hollywood}, has more than $18,000$ video-sentence pairs. We follow the standard evaluation protocol from~\cite{gao2017tall} and use two separate \textit{training} and \textit{testing} splits.

 \noindent \textbf{Standard Evaluation.}
We adopt the same
metrics of Moment-DETR~\cite{lei2021detecting, moon2023query, lee2025bam} for evaluation. 
Specifically, for Moment Retrieval, we report 
the mean average precision (mAP) at Intersection over Union (IoU) thresholds of $0.5$ and $0.75$, as well as the average mAP over IoU thresholds ranging from 0.5 to 0.95 with a step size of $0.05$. We also include Recall@1 (R@1) at IoU thresholds of $0.5$ and $0.75$. 
For a fair comparison, we also include the \textit{test} split of the QVHighlights dataset, which is evaluated on the CodaLab competition platform~\cite{codalab_competitions_JMLR}. Notably, performance at higher IoU thresholds, such as $0.7$, serves as an indicator of more 
precise alignments between 
predicted moments 
and the ground truth.

\input{table/table_spurious_results}

\noindent \textbf{Spurious Evaluation.}
To verify the spurious correlation with overly association the text with background, 
we replace the target moments in video content with random-valued masks while keeping the video duration unchanged, making the masked region the most unrelated moments to text queries.
We then assess the model’s performance under these conditions and report the spurious metric of \emph{Spurious R@1} and \emph{Spurious mAP}, analogous to the standard R@1 and mAP. In this case, lower values are better, as they indicate less spurious correlation.


\noindent \textbf{Implementation Details.}
Our method is implemented in PyTorch~\cite{paszke2019pytorch}. We follow the implementation of QD-DETR~\cite{moon2023query}. For all datasets, we use video features both extracted from SlowFast~\cite{feichtenhofer2019slowfast} pre-trained on Kinetics~\cite{carreira2017quo} and pre-trained CLIP~\cite{radford2021learning} vision encoder, and text feature extracted from pre-trained CLIP~\cite{radford2021learning} text encoder, following Moment-DETR.
\begin{itemize}
    \item For \emph{QVHighlights} dataset, we set batch size to $32$ and an initial learning rate of $1e-4$ with  weight decay of $1e-4$. We set the hidden size $d=256$, layers of encoder/decoder $T=3$, and moment queries $N=10$. The model is trained for 200 epochs. 
    \item For \emph{Charades-STA}, we set batch size to $32$ and use an initial learning rate of $1e-4$ with weight decay of $1e-4$. We set the hidden size $d=256$, layers of encoder/decoder $T=3$, and moment queries $N=10$. The model is trained for 100 epochs and the learning rate is decayed to $\frac{1}{10}$ every $40$ epochs. 
\end{itemize}
\begin{figure*}[t]
    \centering
    \includegraphics[width=1\linewidth]{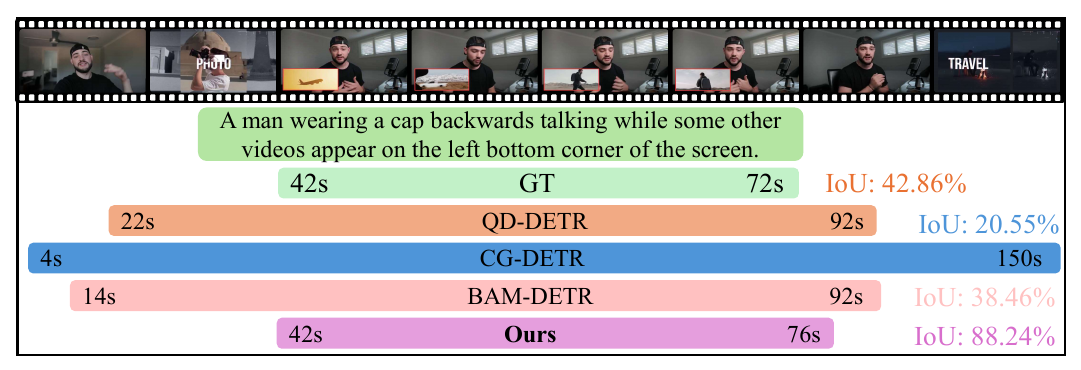}
    \caption{Visualization of results comparison between the baseline models and our \emph{TD-DETR}. We display the relative position of ground truth and prediction moments in colourful boxes, where the green with GT stands for ground truth moment, the orange box with QD-DETR stands for our baseline predictions and the purple box with ours stands for our \emph{TD-DETR}. To quantify the quality of predictions, we mark the start and end timestamps in the box and mark the IoU with the same colour as the models. Best read in colour.}
    \label{fig:Qualitative_Analysis}
\end{figure*}
We use mini-batch and AdamW~\cite{loshchilov2017decoupled} gradient descent algorithm to optimize the network parameters, initialize weights with Xavier init~\cite{glorot2010understanding}, use dropout of $0.1$ for transformer layers and $0.5$ for input projection layers and set cost coefficient as the same as loss balancing coefficient on all datasets. For Dynamic Video Synthesizer, the selection Probability $\alpha$ is set to $0.5$ and for Dynamic Temporal Identifier the $\beta$ is set to $0.7$ on all datasets.
The total training time is approximately 8 hours on \emph{QVHighlight} and 6 hours for \emph{Charades-STA}, with a single NVIDIA RTX 2080 GPU.

\input{table/table_ablation}


\subsection{Results on Standard Evaluation}
As illustrated in Table~\ref{table::QVHighlight} and Table~\ref{table::charades}, we compare \emph{TD-DETR} with several baseline methods on QVHighlights and Charades-STA benchmarks. 

\noindent \textbf{Results on QVHighlights.}
On \emph{QVHighlights} \textit{test} split, we compare our model against proposal-based, proposal-free, and DETR-based methods, as shown in Table~\ref{table::QVHighlight}. As observed, our \emph{TD-DETR} achieves state-of-the-art performance across all evaluation metrics. Notably, \emph{TD-DETR} shows remarkable improvements compared to the other methods. Performance at higher IoU thresholds (\eg, 0.7) provides a clearer indication of the alignment between predicted moments and the ground truth. Our \emph{TD-DETR} outperforms the previous state-of-the-art model by a substantial margin, with improvements of up to $3.56\%$ in R@1@0.7 and $2.14\%$ in mAP@0.75.

\noindent \textbf{Results on Charades-STA.}
We evaluate our method methods on \emph{Charades-STA} \textit{test} split, compared with proposal-based, proposal-free and DETR-based methods. As shown in Table~\ref{table::charades}, \emph{TD-DETR} achieves state-of-the-art performance across all evaluation metrics. Notably, our method achieves improvements of up to $2.46\%$ in R1@0.7 and $1.57\%$ in R1@0.5 on this challenging benchmark. Recall that performance at high IoU thresholds stands for the precision with which predictions match the ground truth data.

\subsection{Results on Spurious Correlation Evaluation}
Moment retrieval models are often prone to inaccurate predictions due to spurious correlations, a typical example being when predictions rely overly on background cues. To investigate this issue, we adopt the spurious evaluation setting pre-defined in Sec~\ref{sec:set_up}. 

We evaluate our methods on several strong baselines, including QD-DETR~\cite{moon2023query}, CG-DETR~\cite{moon2023correlation} and BAM-DETR~\cite{lee2025bam}.
The results in Table~\ref{table::spurious_correlation} demonstrate the alleviation of spurious correlation issue, as indicated by the improved \emph{Spurious R1} and \emph{Spurious mAP} across all compared baselines while boosting the standard mAP. For instance, our methods improve QD-DETR~\cite{moon2023query} in 
Spurious R1@0.9 from $10.40$ to $8.15$ ($30.43\%$), 
with 
standard mAP@0.75 increasing from $41.22$ to $49.05$ ($18.99\%$). 
While for BAM-DETR~\cite{lee2025bam}, it achieves the most substantial improvement, improving Spurious R1 to $1.61$ and Spurious mAP to $1.73$. The results on spurious correlation evaluation demonstrate the effectiveness of our proposed approaches to alleviate the spurious correlation issue and also the generalization across different baselines.

\subsection{Qualitative Analysis}

In this subsection, we analyze the impact of mitigating spurious correlations in our \emph{TD-DETR} model.
As illustrated in Figure~\ref{fig:Qualitative_Analysis}, the entire video depicts a man speaking to the camera, while the target moment is described as: ``A man wearing a cap backwards talking while some other videos appear in the bottom left corner of the screen."
Notably, baseline models fail to distinguish between ``The man is talking" and ``The man is talking while other videos appear in the corner," indicating that background elements are overly correlated with the textual query. In contrast, our \emph{TD-DETR} accurately predicts the target moment without being misled by background frames, demonstrating its ability to associate the information between text and the proper frames rather than relying on spurious correlations.

\subsection{Ablation Studies}

We conduct ablation studies on both QVHighlights and Charades-STA benchmarks to validate the effectiveness of each proposed component. \emph{VSDC} and \emph{TDEM} refer to Video Synthesizer for Dynamic Context and Temporal Dynamics Enhancement Module, respectively.
As illustrated in Table~\ref{table::ablation}, rows (b) to (c) show the effectiveness of each component compared to the baseline model (a), and (d) demonstrate the overall effectiveness of all the components. In detail, on QVHighlight, Video Synthesizer for Dynamic Context contributes improvement of $5.60\%$ in R1@0.7, $9.18\%$ in mAP@0.75 and $9.05\%$ in average mAP while Temporal Dynamics Enhancement Module contributes improvement of $3.16\%$ in R1@0.7, $4.19\%$ in mAP@0.75 and $5.41\%$ in average mAP for Moment Retrieval.
With all components integrated, we observe a substantial $14.75\%$, $14.20\%$ and $15.30\%$ improvement in the R1@0.7, mAP@0.75 and average mAP, respectively. Moreover, the spurious correlation is also improved, where $30.43\%$ in Spurious R1@0.9 and $21.63\%$ in Spurious mAP. The same phenomenon can be observed in Charades-STA as well. 
More experiments and analyses are included in the supplementary materials.

\subsection{Generalization across Different Architectures.}
We validate the generalization by incorporating our proposed approaches into different architectures. SnAG~\cite{mu2024snag} is a hybrid framework, which integrates transformer for feature fusion and CNN for prediction, while QD-DETR~\cite{moon2023query} and BAM-DETR~\cite{lee2025bam} follow DETR-like architecture. 
\input{table/table_generalization}

As illustrated in Table~\ref{tab:cross_different_archtectures}, for the hybrid architecture, our method via SnAG achieves improvements of up to $10.04\%$ in R1@0.7 and $9.46\%$ in mAP on QVHighlights and $6.73\%$ in R1@0.5 and $13.48\%$ in R1@0.7 on Charades-STA. For DETR-like architecture, our approach enhances QD-DETR with significant gains of $19.23\%$ in R1@0.7 and $18.87\%$ in mAP on QVHighlights and $6.25\%$ in R1@0.5 and $23.96\%$ in R1@0.7 on Charades-STA.
Our strategies significantly improve the performance of both hybrid and DETR-like baselines on both benchmarks, demonstrating the generalization of our approach.

%% file: table/table_QVH_result.tex
\begin{table*}[t]
    \setlength{\tabcolsep}{12pt} 
	\centering
	{\normalsize 
    \begin{threeparttable}
        \begin{tabular}{l||cc|ccc}
        \hlineB{2.5}
        \multicolumn{1}{c||}{\multirow{2}{*}{Method}}                        & \multicolumn{2}{c|}{MR-R1}          & \multicolumn{3}{c}{MR-mAP}    \\ \cline{2-6} 
        \multicolumn{1}{c||}{}                                        & @0.5           & @0.7           & @0.5           & @0.75          & Average      \\ \hlineB{2.5}
        MCN~\cite{anne2017localizing}\scriptsize{~\textit{ICCV'17}}                & 11.41          & 2.72           & 24.94          & 8.22           & 10.67     \\
        CAL~\cite{escorcia2019finding}\scriptsize{~\textit{arXiv'19}}              & 25.49          & 11.54          & 23.40          & 7.65           & 9.89      \\
        XML~\cite{lei2020tvr}\scriptsize{~\textit{ECCV'20}}                        & 41.83          & 30.35          & 44.63          & 31.73          & 32.14     \\
        XML+~\cite{lei2021detecting}\scriptsize{~\textit{NIPS'21}}                 & 46.69          & 33.46          & 47.89          & 34.67          & 34.90     \\
        SnAG\tnote{\dag}~~\cite{mu2024snag}\scriptsize{~\textit{CVPR'24}}          & 59.79          & 48.10          & 58.63          & 44.37          & 42.71     \\
        \emph{SnAG /w TD-DETR}                                        & \textbf{66.48}          & \textbf{52.93}          & \textbf{63.71}          & \textbf{49.11}          & \textbf{46.75}     \\ \hline
        Moment-DETR~\cite{lei2021detecting}\scriptsize{~\textit{NIPS'21}}          & 52.89          & 33.02          & 54.82          & 29.40          & 30.73     \\
        UMT~\cite{liu2022umt}\scriptsize{~\textit{CVPR'22}}                        & 56.23          & 41.18          & 53.83          & 37.01          & 36.12     \\
        MomentDiff~\cite{li2023momentdiff}\scriptsize{~\textit{NIPS'23}}           & 57.42          & 39.66          & 54.02          & 35.73          & 35.95     \\
        QD-DETR~\cite{moon2023query}\scriptsize{~\textit{CVPR'23}}                 & 62.40          & 44.98          & 62.52          & 39.88          & 39.86     \\
        UniVTG~\cite{lin2023univtg}\scriptsize{~\textit{ICCV'23}}                  & 58.86          & 40.86          & 57.60          & 35.59          & 35.47     \\
        CG-DETR\cite{moon2023correlation}\scriptsize{~\textit{arXiv'23}}            & 65.40          & 48.40          & 64.50          & 42.80          & 42.90     \\
        UVCOM~\cite{xiao2024bridging}\scriptsize{~\textit{CVPR'24}}                & 63.55          & 48.70          & 64.47          & 44.01          & 43.27     \\
        BAM-DETR\cite{lee2025bam}\scriptsize{~\textit{ECCV'24}}                    & 64.53          & 48.64          & 64.57          & 46.33          & 45.36     \\
        \emph{TD-DETR} (\textbf{Ours})               & \textbf{64.53}$_{\pm_{0.62}}$          & \textbf{50.37}$_{\pm_{0.53}}$          & \textbf{66.21}$_{\pm_{0.21}}$           & \textbf{47.32}$_{\pm_{0.53}}$          & \textbf{46.69}$_{\pm_{0.26}}$  \\ \hlineB{2.5}
        \end{tabular}
        \tnote{\dag}reproduced by the official code
        
	\caption{Performance comparison on QVHighlights \textit{test} split. Our experimental results are averaged over three runs and `$\pm$' denotes the standard deviation. For the compared methods, the results are copied from their original papers and we reproduce SnAG by the official code via slowfast+clip features.}
	\label{table::QVHighlight}
    \end{threeparttable}
        }
\end{table*}

%% file: table/table_charades_results.tex
\begingroup
\setlength{\tabcolsep}{14pt} 
\renewcommand{\arraystretch}{1} 
\begin{table}
    \centering
    {
    \normalsize
    \begin{threeparttable}
        \begin{tabular}{l||cc}
        \hlineB{2.5}
        Method & R1@0.5 & R1@0.7 \\ \hlineB{2.5}
        CAL~\cite{escorcia2019finding} & 44.90 & 24.37 \\
        2D TAN~\cite{zhang2020learning} & 39.70 & 23.31 \\
        VSLNet~\cite{zhang2020span} & 47.31 & 30.19 \\
        IVG-DCL~\cite{nan2021interventional} & 50.24 & 32.88 \\
        SnAG\tnote{\dag}~~\cite{mu2024snag} & 65.72 & 37.32 \\
        \emph{SnAG /w TD-DETR} & \textbf{70.14} & \textbf{42.35} \\ \hlineB{2.5}
        Moment-DETR~\cite{lei2021detecting} & 53.63 & 31.37 \\
        Moment-Diff~\cite{li2023momentdiff} & 55.57 & 32.42 \\
        UMT~\cite{liu2022umt} & 48.31 & 29.25 \\
        QD-DETR~\cite{moon2023query} & 57.31 & 32.55 \\ 
        CG-DETR\cite{moon2023correlation} & 58.40 & 36.30 \\
        BAM-DETR\cite{lee2025bam} & 59.95 & 39.38 \\
        \emph{TD-DETR} (\textbf{Ours}) & \textbf{60.89} & \textbf{40.35} \\  \hlineB{2.5}
        \end{tabular}
        \tnote{\dag}reproduced by the official code
    \end{threeparttable}
    \caption{Performance comparison on Charades-STA \textit{test} split. Our experimental results are averaged over three runs. For all the compared methods, the results are taken from
their original papers.}
    \label{table::charades}
    }
\end{table}
\endgroup

%% file: table/table_spurious_results.tex
\begin{table}
        \setlength{\tabcolsep}{3pt} 
	\centering
	{\small
        \begin{tabular}{c||cc|cc|cc}
        \hlineB{2.5}
        \multicolumn{1}{c||}{\multirow{2}{*}{Method}} & \multicolumn{2}{c|}{Spurious R1 \blue{$\bm{\downarrow}$}}          & \multicolumn{2}{c|}{Spurious mAP \blue{$\bm{\downarrow}$}} & \multicolumn{2}{c}{Standard mAP \red{$\bm{\uparrow}$}} \\ \cline{2-7} 
        
        \multicolumn{1}{c||}{}  & @0.7      &@0.9      & @0.75    & Avg.    & @0.75    & Avg.   \\ \hline

        QD-DETR & 9.35 & 5.29 & 9.90 & 10.40 & 41.82 & 41.22 \\
        \rowcolor[gray]{0.9}
        Ours w/ QD & 8.26 & 3.68 & 7.46 & 8.15 & 49.86 & 49.05 \\
        CG-DETR & 4.65 & 1.29 & 5.55 & 6.14 & 45.70 & 44.90 \\
        \rowcolor[gray]{0.9}
        Ours w/ CG & 2.58 & 0.39 & 3.38 & 4.41 & 49.16 & 48.38 \\
        BAM-DETR  & 7.16 & 1.87 & 6.30 & 6.72 & 48.56 & 47.61 \\
        \rowcolor[gray]{0.9}
        Ours w/ BAM & 1.61 & 0.52 & 1.73 & 1.98 & 49.62 & 48.67  \\ 
        \hlineB{2.5}
        \end{tabular}
        \caption{Performance comparison on QVHighlights \textit{val} split with random masks. The target clips in the video content are replaced with random-valued masks while keeping the video duration unchanged. 
        We compare the baseline models—QD-DETR, CG-DETR, and BAM-DETR—against their respective versions integrated with our proposed \emph{TD-DETR}, marked \emph{Ours w/}, under the same experimental settings. The metrics here have been explained in section~\ref{sec:set_up}.
	  \label{table::spurious_correlation}
        }
        }
\end{table}

%% file: table/table_ablation.tex
\begin{table*}[t]
    \setlength{\tabcolsep}{5pt} 
	\centering
	{\small
        \begin{tabular}{c|c|c||cc|ccc|cc|cc|cc|cc}
        \hlineB{2.5}
        \multicolumn{1}{c|}{\multirow{3}{*}{}} & \multicolumn{1}{c|}{\multirow{3}{*}{VSDC}} & \multicolumn{1}{c||}{\multirow{3}{*}{TDEM}} & \multicolumn{9}{c|}{QVHighlight} & \multicolumn{4}{c}{Charades-STA} \\ \cline{4-16} 
        \multicolumn{1}{c|}{}                  & \multicolumn{1}{c|}{}                      & \multicolumn{1}{c||}{}                      & \multicolumn{2}{c}{R1\red{$\bm{\uparrow}$}} & \multicolumn{3}{c|}{mAP\red{$\bm{\uparrow}$}} & \multicolumn{2}{c}{Spurious R1\blue{$\bm{\downarrow}$}} & \multicolumn{2}{c|}{Spurious mAP\blue{$\bm{\downarrow}$}}  & \multicolumn{2}{c|}{R1\red{$\bm{\uparrow}$}} & \multicolumn{2}{c}{Spurious R1\blue{$\bm{\downarrow}$}} \\ \cline{4-16} 
        
        \multicolumn{1}{c|}{} &\multicolumn{1}{c|}{} & \multicolumn{1}{c||}{}  & @0.5           & @0.7           & @0.5           & @0.75          & Avg.           & @0.7           & @0.9           & @0.75          & Avg.           & @0.5              & @0.7           & @0.7              & @0.9\\ \hline

        (a) &  &  & 61.12 & 46.77 & 62.45 & 43.66 & 42.54 & 9.35 & 5.29 & 9.90 & 10.40 & 57.31 & 32.55 & 25.72 & 6.31 \\
        (b) &\ding{51} & & 63.47 & 49.39 & 64.82 & 47.67 & 46.39 & 8.77 & 3.87 & 8.64 & 8.91 & 39.12 & 63.67 & 23.15 & 5.42 \\ 
        (c) & & \ding{51} & 62.93 & 48.25 & 64.22 & 45.49 & 44.84 & 8.84 & 4.0 & 9.10 & 9.56 & 38.51 & 60.80 & 24.03 & 5.73 \\
        (d) & \ding{51} & \ding{51} & \textbf{65.88} & \textbf{53.67} & \textbf{66.43} & \textbf{49.86} & \textbf{49.05} & \textbf{8.26} & \textbf{3.68} & \textbf{7.46} & \textbf{8.15} & \textbf{60.89} & \textbf{40.35} & \textbf{22.13} & \textbf{4.82} \\ \hlineB{2.5}
        \end{tabular}
	\caption{Ablation studies on QVHighlights $\textit{val}$ split and Charades-STA $\textit{test}$ split. \emph{VSDC} and \emph{TDEM} stand for Video Synthesizer for Dynamic Context and Temporal Dynamics Enhancement Module respectively. All the ablation results are averaged over three runs. We also reveal spurious metrics of the proposed modules.}
	\label{table::ablation}
        }
\end{table*}

%% file: table/table_generalization.tex
\begin{table}[ht]
\vspace{-0.36cm}
\footnotesize
\centering 
\setlength{\tabcolsep}{0.56mm}
\begin{tabular}{c||ccc||cc}
  \hlineB{2.5}
   \multirow{2}{*}{Method} & \multicolumn{3}{c||}{QVHighlights val} & \multicolumn{2}{c}{Charades-STA test} \\
   \cline{2-6}
   & R1@0.7 & mAP@0.75 & mAP & R1@0.5 & R1@0.7 \\
   \hlineB{2.5}
    \hline
    CG & 52.10 & 45.70 & 44.90 & 58.40 & 36.30  \\
    \rowcolor[gray]{0.9}
    Ours w/ CG & 53.25\color{red}{$_{+{\textbf{\textit{1.15}}}}$} & 49.16\color{red}{$_{+{\textbf{\textit{3.46}}}}$} & 48.38\color{red}{$_{+{\textbf{\textit{3.48}}}}$} & 59.35\color{red}{$_{+{\textbf{\textit{0.95}}}}$} & 37.84\color{red}{$_{+{\textbf{\textit{1.54}}}}$}\\
    BAM & 51.61 & 48.56 & 47.61 & 59.95 & 39.38 \\
    \rowcolor[gray]{0.9}
    Ours w/BAM & 52.87\color{red}{$_{+{\textbf{\textit{1.26}}}}$} & 49.62\color{red}{$_{+{\textbf{\textit{1.06}}}}$} & 48.82\color{red}{$_{+{\textbf{\textit{1.21}}}}$} & 60.92\color{red}{$_{+{\textbf{\textit{0.97}}}}$} & 40.25\color{red}{$_{+{\textbf{\textit{0.87}}}}$} \\
    QD & 46.66 & 41.82 & 41.22 & 57.31 & 32.55  \\
    \rowcolor[gray]{0.9}
    Ours w/ QD & 53.67\color{red}{$_{+{\textbf{\textit{7.01}}}}$} & 49.86\color{red}{$_{+{\textbf{\textit{8.04}}}}$} & 49.00\color{red}{$_{+{\textbf{\textit{7.78}}}}$} & 60.89\color{red}{$_{+{\textbf{\textit{3.58}}}}$} & 40.35\color{red}{$_{+{\textbf{\textit{7.80}}}}$} \\
    \hline
    SnAG & 48.10  & 44.37 & 42.71 & 65.72 & 37.32\\
    \rowcolor[gray]{0.9}
    Ours w/ SnAG & 52.93\color{red}{$_{+{\textbf{\textit{4.83}}}}$} & 49.11\color{red}{$_{+{\textbf{\textit{4.74}}}}$} & 46.75\color{red}{$_{+{\textbf{\textit{4.04}}}}$} & 70.14\color{red}{$_{+{\textbf{\textit{4.42}}}}$} & 42.35\color{red}{$_{+{\textbf{\textit{5.03}}}}$} \\ 
  \hlineB{2.5}
\end{tabular}
\caption{Generalization across different architectures on both QVHighlight \textit{val} split and Charades-STA \textit{test} split. CG stands for CG-DETR, while QD stands for QD-DETR and BAM stands for BAM-DETR.}
\label{tab:cross_different_archtectures}
\end{table}

%% file: sec/X_supple.tex
\clearpage
\setcounter{page}{1}
\setcounter{section}{0}
\setcounter{subsection}{0}
\setcounter{table}{0}
\setcounter{figure}{0}
\definecolor{citeblue}{HTML}{357cbc}
\renewcommand{\thefigure}{S\arabic{figure}}
\renewcommand{\thetable}{S\arabic{table}}
\renewcommand{\thesection}{S\arabic{section}}

\maketitlesupplementary
\section{Details of Network and Objectives}
\label{sec:formulation}
In this section, we present our network and loss functions in detail.

\noindent \textbf{Transformer encoder-decoder with prediction heads.} We follow the architectural principles outlined in~\cite{lei2021detecting, moon2023query} for the design of our transformer and prediction heads, with modifications introduced in the encoder. Specifically, we integrate our proposed \emph{Temporal Dynamic Tokenizer} into the encoder to address spurious correlations effectively.

Given a spurious pair $p_i' = \{\tilde{V}'_i, \tilde{V}'_{k_i}\}$, this module processes all video pairs uniformly. 
As introduced in Section~\textcolor{citeblue}{3.2}, we use $T$ to denote the temporal dynamics. To incorporate these dynamics, we employ two transformer encoder layers with cross attention that facilitate bidirectional interactions: (1) between the temporal dynamics and the text query, and (2) between the video content and the text query. Once the temporal dynamics and video content are individually aligned with the text, we apply a weighted element-wise addition to combine their outputs. This rejected representation is subsequently processed through a standard transformer layer to refine the contextual understanding.
Given a spurious pair $p_i = \{\tilde{V}_i, \tilde{V}_{k_i}\}$, our approach generates a refined spurious pair $p_i' = \{\tilde{V}'_i, \tilde{V}'_{k_i}\}$ that incorporates these temporal and semantic enhancements.

\noindent \textbf{Loss Functions.}
We compute the loss between the predicted output $\hat{y}$ and its corresponding ground truth $y$ ($m_i$) for $\tilde{V}''_i$, as well as between $\hat{y}'$ and its ground truth $y'$ ($m'_i$) for $\tilde{V}''_{k_i}$. The predictions are matched with their targets, and the loss is calculated using L1 loss, generalized IoU (gIoU) loss, and cross-entropy loss, respectively, as described in~\cite{lei2021detecting}.

\section{Sensitiveness Analysis} 
\subsection{Video Synthesizer for Dynamic Context}
\label{sec:hyperparameter}

In Section \textcolor{citeblue}{3.1}, we construct a new sample $\tilde{V}_{k_i}$ with dynamic context from spurious pair $p_i = \{V_i, V_{k_i}\}$ as follows, 
\begin{equation}
    \tilde{V}_{k_i} = \alpha \cdot V_i + (1-\alpha) \cdot V_{k_i},
\end{equation}
where $\alpha$ represents the sampling ratio of $V_i$ while $1-\alpha$ corresponds to $V_{k_i}$.

We examine the impact of the sampling ratio $\alpha$ on the quality of the synthesized samples. In detail, we adapt $\alpha$ ranging from $0.1$ to $0.9$ with a step size of $0.2$. 

As illustrated in Table~\ref{table::ablation_alpha}, when the sampling ratio $\alpha$ increases, the synthesized video incorporates more tokens from the videos containing the target moments with corresponding dynamic contexts, thus improving the performance of moment retrieval.
The performance starts to decline from $\alpha = 0.9$, due to the lack of dynamics of the contexts.
Specifically, when $\alpha = 1.0$, the synthesized video is identical to the original video. 
This ablation study on $\alpha$ demonstrates the effectiveness of our \emph{Video Synthesizer for Dynamic Context} in improving model performance by balancing contextual information and target moment focus. Besides, even with various sampling ratios $\alpha$, our method still achieves promising results, which demonstrate the robustness of the proposed method.
\input{table/table_ablation_alpha}
\input{table/table_ablation_beta}
\input{table/table_InternVideo}

\subsection{Dynamics Enhancement}

In section \textcolor{citeblue}{3.2}, the model learns from both dynamic and video information via cross-attention machines. To emphasize the learned dynamic information, we inject text-guided dynamic representation $T'$ into video $\tilde{V}_{i}$ as follows,
\begin{equation}
    \tilde{V}'_{i} = \beta \cdot \tilde{V}_{i} + (1-\beta) \cdot V_{i},
\end{equation}
where $\beta$ represents the injection ratio of the video information we used, while $1-\beta$ corresponds to temporal information $T'$.
\begin{figure}[ht]
    \centering
    \includegraphics[width=1\linewidth]{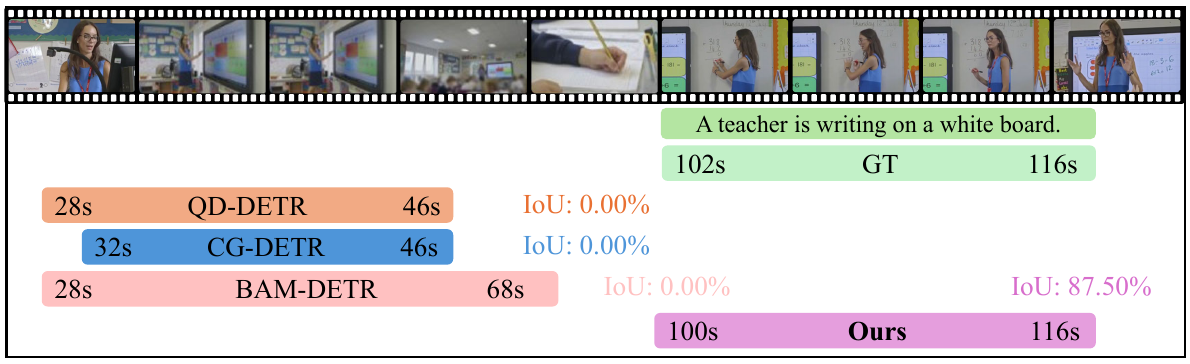}
    \caption{Model prediction on query ``A teacher is writing on a whiteboard.". Baselines tend to predict the teacher writing on a screen instead of the target moment which indicates baseline models fail to distinguish between ``screen" and ``whiteboard".}
    \label{fig:extend_fig1}
\end{figure}
We also examine the impact of the injection ratio $\beta$ on the quality of the injected videos. In detail, we adapt $\beta$ ranging from $0.3$ to $0.9$ with a step size of $0.2$ and evaluate $\beta=0$ as an extra experiment. 
As illustrated in Table~\ref{table::ablation_beta}, when the injection ratio $\beta$ decreases, the video is injected with more temporal information, thus improving the performance of moment retrieval.
The performance achieves the highest performance when $\beta=0.7$, which indicates the benefits of dynamic enhancement.
Specifically, when $\beta=1.0$, no dynamic information is injected into the video, thus the performance drops a lot in contrast to those with dynamics representation.
Note that when $\beta=0.0$, the model relies solely on temporal dynamic information, which leads to poor predictions due to the absence of any object-related cues.
This ablation study on $\beta$ validates the effectiveness of our \emph{Temporal Dynamics Enhancement} in boosting moment retrieval by encouraging our model to align text queries with temporal-dynamic representations. 
Besides, even with various sampling ratios $\beta$, our method still achieves promising results, which demonstrate the robustness of the proposed method.
\input{table/table_similarity}

\section{Ablation Analysis on Video Sampling Strategy}
In Section~\ref{sec::VSDC}, we select a video that is contextually similar to \( V \) to ensure both the challenge and rationality of the synthesized video. As shown in Table~\ref{tab:sim}, we compare our similarity-based selection strategy with a random sampling approach on QVHighlights and Charades-STA. The \textit{w/ random} selection still outperforms the QD-DETR baseline but falls short of \emph{w/ similarity}, demonstrating the effectiveness of our approach in generating meaningful and challenging synthetic video contexts.

\section{Additional Results of Predicted Results}
\subsection{More Prediction Examples.}
\begin{figure}[ht]
    \centering
    \includegraphics[width=1\linewidth]{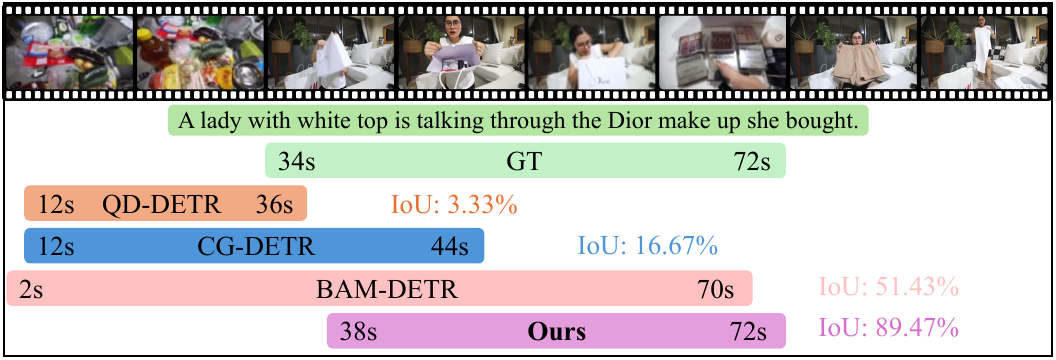}
    \caption{Model prediction on query ``A lady with white top is talking through the Dior make-up she bought.". Baselines tend to predict the woman with some food and clothes instead of the target moment which indicates baseline models fail to distinguish between ``make-up" and ``clothes".}
    \label{fig:extend_fig2}
\end{figure}

More visualization results of predictions and baselines comparison from our proposed \emph{TD-DETR} model are presented in Figure~\ref{fig:extend_fig1} and Figure~\ref{fig:extend_fig2}.

\subsection{New Validation Split on Spurious Correlation}
Except for Spurious R@1 and Spurious mAP, we introduce a new validation split based on the QVHighlight validation set to further evaluate spurious correlations. Specifically, similar to Section~\ref{sec::VSDC}, we replace the contextual frames of a video with clips from another video, creating a more dynamic and diverse context. This modification aims to disrupt excessive contextual associations and better assess the model’s robustness against spurious correlations.
The modified validation split will be released publicly with our code.

All illustrated in Table~\ref{tab:new_val}, our proposed \emph{TD-DETR} still achieves state-of-the-art performance among all baselines on such a dynamic context validation.

\input{table/table_new_spurious_val}

\section{Generalization across Different Feature Representations}
With the rapid advancement of large multi-modal models in video understanding, InterVideo2—a video foundation model introduced by~\cite{wang2024internvideo2}—has demonstrated strong capabilities in moment retrieval. Beyond SlowFast~\cite{feichtenhofer2019slowfast}, we further evaluate our model’s generalization across different feature representations. All illustrate in Table~\ref{tab:cross_different_feature}, the proposed \emph{TD-DETR} also achieve state-of-the-art performance. Our TD-DETR outperforms the previous state-of-the-art model by a substantial margin, with improvements of up to $3.81\%$ in R@1@0.7 and $3.88\%$ in mAP@0.75 on QVHighlights \textit{val} split and $7.30\%$ in R@1@0.5 and $9.68\%$ in R1@0.7 on Charades-STA \textit{test} split.

%% file: table/table_ablation_alpha.tex
\begin{table}[ht]
        \setlength{\tabcolsep}{6pt} 
	\centering
	{\small 
        \begin{tabular}{l||cc|ccc}
        \hlineB{2.5}
        \multicolumn{1}{c||}{\multirow{2}{*}{$\alpha$}}                        & \multicolumn{2}{c|}{MR-R1}          & \multicolumn{3}{c}{MR-mAP}                         \\ \cline{2-6} 
        \multicolumn{1}{c||}{}                        & @0.5           & @0.7           & @0.5           & @0.75          & Avg.           \\ \hlineB{2.5}
        0.0          & 11.61          & 3.35           & 23.93          & 7.5            & 10.09                 \\
        0.3          & 65.10          & 51.94          & 65.77          & 48.13          & 47.55                 \\
        0.5          & 64.77          & 51.10          & 66.79          & 49.08          & 47.95                \\
        \rowcolor[gray]{0.9}
        0.7          & 65.88          & 53.67          & 66.43          & 49.86          & 49.05                 \\
        0.9          & 64.19          & 51.23          & 66.29          & 48.88          & 47.94                 \\\hlineB{2.5}
        \end{tabular}
	  \caption{Sensitiveness analysis of sampling ratio $\alpha$ on QVHighlights $\textit{val}$ split.}
        \label{table::ablation_alpha}
        }        
\end{table}

%% file: table/table_ablation_beta.tex
\begin{table}[ht]
        \setlength{\tabcolsep}{6pt} 
	\centering
	{\small 
        \begin{tabular}{l||cc|ccc}
        \hlineB{2.5}
        \multicolumn{1}{c||}{\multirow{2}{*}{$\beta$}}                        & \multicolumn{2}{c|}{MR-R1}          & \multicolumn{3}{c}{MR-mAP}                         \\ \cline{2-6} 
        \multicolumn{1}{c||}{}                        & @0.5           & @0.7           & @0.5           & @0.75          & Avg.           \\ \hlineB{2.5}
        0.1          & 65.10          & 51.94          & 67.37          & 50.12          & 48.87                 \\
        0.3          & 64.77          & 51.48          & 66.50          & 49.76          & 48.48                 \\
        0.5           & 65.15          & 51.26          & 66.24          & 48.44          & 47.81                \\
        \rowcolor[gray]{0.9}
        0.7          & 65.88          & 53.67          & 66.43          & 49.86          & 49.05                 \\
        0.9         & 62.97          & 50.19          & 65.81          & 48.76           & 47.83                 \\\hlineB{2.5}
        \end{tabular}
        \caption{Sensitiveness analysis of sampling ratio $\beta$ on QVHighlights $\textit{val}$ split.}
	  \label{table::ablation_beta}
        }
\end{table}

%% file: table/table_InternVideo.tex
\begin{table*}[ht]
\vspace{-0.36cm}
\footnotesize
\centering 
\setlength{\tabcolsep}{8pt}
\begin{tabular}{c||ccccc||cc}
   \hlineB{2.5}
   \multirow{2}{*}{Method} & \multicolumn{5}{c||}{QVHighlights val} & \multicolumn{2}{c}{Charades-STA test} \\
   \cline{2-8}
    & R1@0.5 & R1@0.7 & mAP@0.5 & mAP@0.75 & mAP & R1@0.5 & R1@0.7 \\
    \hline
    QD-DETR & 68.58 & 52.13 & 67.87 & 45.94 & 45.40 & 66.63 & 42.78 \\
    CG-DETR & 70.27 & 55.62 & 69.17 & 52.62 & 50.93 & 69.11 & 46.13 \\
    BAM-DETR & 69.72 & 55.13 & 69.38 & 52.89 & 51.13 & 68.49 & 48.33 \\
    \emph{TD-DETR} \textbf{(ours)} & \textbf{71.29} & \textbf{57.23} & \textbf{72.99} & \textbf{54.94} & \textbf{53.23} & \textbf{73.49} & \textbf{53.01} \\
    \hlineB{2.5}
  \hline
\end{tabular}
\caption{Comparison of models performance on QVHighlights $\textit{val}$ split using InternVideo2 feature representations.}
\label{tab:cross_different_feature}
\end{table*}

%% file: table/table_similarity.tex
\begin{table}[ht]
\caption{Comparisons across different sampling strategies.}
\vspace{-0.36cm}
\footnotesize
\centering 
\setlength{\tabcolsep}{0.7mm}
\begin{tabular}{c||ccc||cc}
  \hline
   \multirow{2}{*}{Method} & \multicolumn{3}{c||}{QVHighlights} & \multicolumn{2}{c}{Charades-STA} \\
   \cline{2-6}
   & R1@0.7 & mAP@0.75 & mAP & R1@0.5 & R1@0.7 \\
   \hline
    baseline & 46.66 & 41.82 & 41.22 & 57.31 & 32.55  \\
    w/ random & 51.29 & 47.82 & 47.56 & 58.66 & 37.98  \\
    w/ similarity & \textbf{53.67} & \textbf{49.86} & \textbf{49.05} & \textbf{60.89} & \textbf{40.35}\\
    \hline
\end{tabular}
\label{tab:sim}
\end{table}

%% file: table/table_new_spurious_val.tex
\begin{table}
    \setlength{\tabcolsep}{4pt} 
	\centering
	{\small
    \begin{tabular}{c||ccc|ccc}
        \hlineB{2.5}
        \multicolumn{1}{c||}{\multirow{2}{*}{Method}} & \multicolumn{3}{c|}{Standard R1 \red{$\bm{\uparrow}$}}          & \multicolumn{3}{c|}{Standard mAP \red{$\bm{\uparrow}$}} \\ \cline{2-7} 
        
        \multicolumn{1}{c||}{}  & @0.5    & @0.7      &mIOU      & @0.5     & @0.75    & Avg.        \\ \hline

        QD-DETR & 58.29 & 39.29 & 52.76 & 57.25 & 34.86 & 34.78\\
        CG-DETR & 62.03 & 43.77 & 56.57 & 59.9 & 38.3 & 38.48 \\
        BAM-DETR & 59.74 & 41.87 & 54.95 & 60.05 & 39.5 & 39.24 \\
        \emph{TD-DETR}  & \textbf{65.77} & \textbf{46.94} & \textbf{59.43} & \textbf{64.5} & \textbf{42.13} & \textbf{42.21}  \\ \hlineB{2.5}
    \end{tabular}
	\caption{Performance comparison on our dynamic context validation split.}
    \label{tab:new_val}
    }
\end{table}